# Study and Observation of the Variations of Accuracies for Handwritten Digits Recognition with Various Hidden Layers and Epochs using Neural Network Algorithm

Md. Abu Bakr Siddique[1*], Mohammad Mahmudur Rahman Khan[2#], Rezoana Bente Arif[1@], and Zahidun Ashrafi[3$]
[1]Dept. of EEE, International University of Business Agriculture and Technology, Dhaka 1230, Bangladesh
[2]Dept. of ECE, Mississippi State University, Mississippi State, MS 39762, USA
[3]Dept. of Statistics, Rajshahi College, Rajshahi 6100, Bangladesh
absiddique@iubat.edu[*], mrk303@msstate.edu[#], rezoana@iubat.edu[@], zimzahidun@gmail.com[$]

*Abstract*—In recent days, Artificial Neural Network (ANN) can be applied to a vast majority of fields including business, medicine, engineering, etc. The most popular areas where ANN is employed nowadays are pattern and sequence recognition, novelty detection, character recognition, regression analysis, speech recognition, image compression, stock market prediction, Electronic nose, security, loan applications, data processing, robotics, and control. The benefits associated with its broad applications leads to increasing popularity of ANN in the era of 21st Century. ANN confers many benefits such as organic learning, nonlinear data processing, fault tolerance, and self-repairing compared to other conventional approaches. The primary objective of this paper is to analyze the influence of the hidden layers of a neural network over the overall performance of the network. To demonstrate this influence, we applied neural network with different layers on the MNIST dataset. Also, another goal is to observe the variations of accuracies of ANN for different numbers of hidden layers and epochs and to compare and contrast among them.

## I. INTRODUCTION

This Artificial Intelligence (AI) is the emulation of human intelligence such as cognitive functions associated with learning, reasoning, self-correction, and perception by machines [1]. The term Artificial Intelligence was introduced first time as a field in the year 1956 at Dartmouth workshop [2, 3]. As the machines become smarter and smarter and adaptation of automation increases by the passage of time, the demand of Artificial Intelligence (AI) is increasing day by day. In recent days, AI is given considerable attention as it is possible to imitate the human capability to learn reasons and make decisions by machines to handle challenging tasks if intelligence is incorporated.

ANN is being considered as one of the top branches of AI as it is a technique of achieving machine intelligence by mimicking the human brain. An ANN is a highly interconnected parallel computing system inspired by the human mind which is massively parallel connected consisting of approximately 100 billion neurons interconnected with each other [4-6]. ANN is a system comprised of an input layer, an output layer and some hidden layers in between [4]. Each layer is a combination of neurons. In ANN each neuron is a function which takes outputs from all the neurons in the preceding layers and spits out a number depending on the images feed in the system. The aggregated neurons in the input layer are activated by the aggregated pixels with different specific grayscale values of the feeding images to the system. The activation of each neuron in the following layers depends on the weighted sums of all the activations of the preceding layer, and a bias value is added to the result. Then the result is composed of an activation function. As bias defines the threshold value to activate the neuron, it allows activation function to shift left or right to fit the result better. In ANN each neuron is portrayed by a small circle called node [7].

To train the neural network, a cost function is defined which compares the absolute output of the system with the desired output to generate an error. Then the signal is back propagated to the system in a repeated fashion to update weights and biases so that the value of cost function is minimized and the network's performance is increased [1, 8]. Backpropagation algorithm exploits stochastic gradient descent to optimize weight and bias values [9-11]. The valuable impacts of the neural network are increasing rapidly in different sectors. Multi-context integrated deep neural network (MCI-DNN) is proposed to improve the location-tracing of the network users [12]. Moreover, long and short-term memory neural network (LSTM NN) is proffered to pinpoint the contamination level of water [13].

One of the central mysteries in the domain of Neural Network is the total number of implemented hidden layers for the best performance. In some cases, the network converges with a minimal number of hidden layers. Conversely, in some cases, the network needs to have an oversized number of hidden layers for the convergence. Therefore, the motivation of this paper is to observe the impact of the hidden layers of a neural network upon the handwritten digits from the Modified National Institute of Standards and Technology (MNIST) dataset [14]. The mathematical model of this neural network algorithm is implemented in python with tensorflow. As 28×28 handwritten digits are taken as inputs, this model has 784 input nodes, in between 5 hidden layers is introduced with 500 nodes for each layer, and output layer consists of 10 nodes, each of them represents digits from 0, 1, 2, 3, 4, 5, 6, 7, 8, and 9 respectively. Finally, accuracies of the network are observed for a different number of hidden layers and iterations, and the comparison is made among them.

## II. LITERATURE REVIEW

ANN has been ruling the various sectors of technology. It is widely momentous these days because of making human life more comfortable by dealing with the machines in a more natural way. From medical science to server security system ANN is strongly imputing its dominance. Apart from the powerful impacts on the other sectors like recognition of different signals in medical science (e.g. automatic detection of spike-wave complexes in EEG-signals, detection of cancer tumors in X-ray photographs etc.), automatic recognition of registration plate numbers of vehicles, identification of the dynamic pattern of signatures, speech recognition, machine condition monitoring, signal source separation, electric power load forecasting, Artificial Neural Network is playing an essential role in handwritten data recognition in machine learning nowadays [15]. The rapid implementation of deep learning in addition to reinforcement learning in bioimaging sector is making the better understanding of the machines. On top of that, it is lessening the expense of enumeration [16]. Another utilization of neural network which is a layered neural network has taken the performance of bioinformatics along with data and image processing at a higher dimension [17]. Research has been done to improve the visualizing concept in the neural network by means of a technique of blurring and de-blurring method which might be beneficial to its (neural network) performance [18]. To minimize the response time in the neural network, method and analysis were proposed. The objective of the proposal was to achieve further accuracy in the performance of the networking system [19]. Using the most popular approaches of the neural network; Deep Neural Network (DNN), Deep Belief Network (DBN) and Convolutional Neural Network (CNN) the problems in handwritten digit recognition can be recovered. However, more perfection, as well as more accuracy, is required to perform this task with minimal error and loss. With the support of Back Propagation (BP), neural network handwritten digit can be recognized [20]. There are several techniques of character recognition, and they all have robust and ailing sides [21]. By developing the accuracy, the weakness in handwritten digit can be minimized. The higher the efficiency, the better the performance of the network will be achieved. In this kind of researches of handwritten data recognition by the neural network, the importance of the MNIST resources are boundless. By using different datasets from MNIST datasets, the accuracy can be checked. To boost the performance of the neural network in order to make apace with the human brain, more accuracy in the performance of ANN is required. One research has shown this accuracy of 89.5% in character recognition [22]. In another past research, using a Multilayer Perceptron (MLP) with one hidden layer of English letters has been attempted to be recognized as having about 70% of accuracy [23]. Again, some methods like Gradient Features and Normalization-Cooperated Gradient Feature were used to increase the performance of the computer in recognizing characters [24]. In previous researches, the error variations were shown for different cases [25], whereas we are dealing with the accuracy in the performance of ANN for the different layers, batch sizes and epochs for handwritten digit recognition. Also, some researches were done to optimize the cost of the ANN [26]. In the past, recognition of mathematical expressions was done though there was no calculation of accuracy or error [27]. In this paper, with the assistance of the MNIST datasets, the accuracies in ANN response or performance for various epochs, different hidden layers and batch size has been developed.

## III. BIOLOGICAL NEURAL NETWORK (BNN) BASICS

The Human brain is a highly sophisticated, nonlinear parallel computing system. The neural system of a human body can be categorized into three stages: i) Receptors, ii) Neural network and iii) Effectors. The Biological neural network is a massively sizeable parallel interconnection of myriad neurons in the brain. The Receptors obtain knowledge from the environments and passes it to the neurons as electrical impulses. The Neural network (NN) processes the electrical signals to make a suitable decision of outputs. Finally, the effectors translate the electrical signals from NN to generate interactions with the environments. Figure 1 demonstrates the bidirectional communication between three stages of the neural system [28].

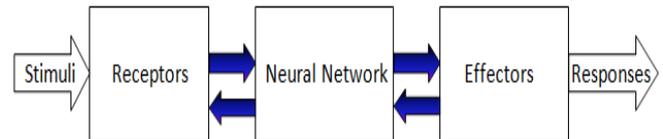

Fig. 1. Stages of Biological Neural System Showing the Steps of Data Processing and Analyzing

A neuron is a primary processing unit in a BNN responsible for receiving, processing and transmitting information via electrical and chemical pulses. Each neuron composed of soma, axon, synapses, and dendrites. Neurons are connected with each other through synaptic junctions. Dendrites receive signal input from surrounding neurons and pass it to axon through soma. Axon of a neuron is connected with the dendrite of another neuron through the synaptic junction. Axon transmits the signal from one neuron to others. The output of the synaptic terminal either incites or impedes the neuron. A neuron fires only if excitatory signals achieve a specific threshold value by exceeding inhibitory signals [29]. Though the processing speed of a neuron ($10^{-3}$ seconds) is much slower than the processing speed of a today's Integrated Circuit ($10^{-9}$ seconds), as billions of neurons exist with trillions of interconnections among them, the elaborate visual perception of human occurs within 100 milliseconds [30, 31]. BNN has very high fault tolerances and can learn from experiences.

## IV. MODELING OF ARTIFICIAL NEURAL NETWORK (ANN)

### A. The Basic Model of an ANN

ANN is a neurocomputing system inspired by the biological modeling of the human brain. In short, an ANN is just a function consists of an input layer, an output layer and several hidden layers in between. The farthest to the left layer in the neural network is addressed as the input layer, and the neurons in this layer are treated as input neurons or input nodes. The farthest to the right layer is addressed as the output layer, and the neurons in this layer are treated as output neurons or output nodes. The middle layers are known as hidden layers. A network may have a single hidden layer or multiple hidden layers or no hidden layer.

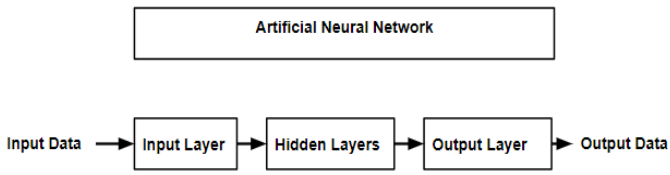

Fig. 2. Necessary steps of ANN to mimic the performance of the biological neural network

An artificial neuron consists of 4 essential elements: i) Nodes ii) Weights, iii) Biases, iv) Activation function. ANN is a multilayered weighted directed function in which each node constitutes an artificial neuron, and directed edges with weights connect neuron inputs and neuron outputs. ANN receives input signals from the outside world in the vector form of images and patterns. Each node in the input layer is activated with a number depending on the types of images feed into the network as input. Then the activation is multiplied by its matching weights. Weight value determines the strength of the connection between neurons in the system. A bias value is added with the weighted sum, and then the result is fed to an activation function. Bias is added to maximize the network's output. A threshold value is configured to circumscribe the response to arrive at the expected value. Activation function activates the output neuron only if the result is higher than the threshold value. Figure 3 shows the basic model of an artificial neural network.

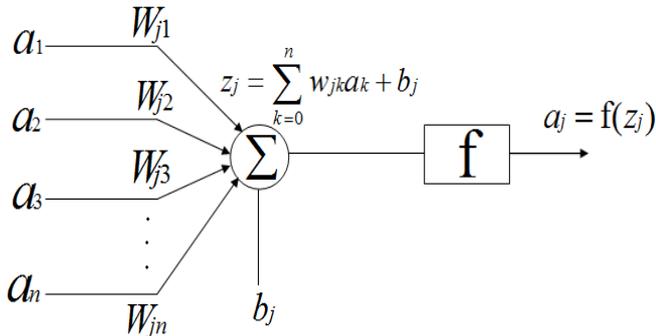

Fig. 3. The basic model of an ANN for simple implementation

In figure 3, the activations of each input neurons are $a_1, a_2, a_3....a_n$, and their corresponding weights $w_{j1}, w_{j2}, w_{j3}....w_{jn}$ connect these activations respectively. A bias $b_j$ is applied to the output neurons. The weighted sum to the output neuron is:

$$z_j = \sum_{k=0}^{n} w_{jk} a_k + b_j \quad ......(1)$$

Then the amount is fed to an activation function, and thus the final output is:

$$a_j = f(z_j) \quad ......(2)$$

### B. A Neural Network Model to Classify Handwritten Digits

To recognize handwritten digits, an eleven-layered feed-forward neural network with one input layer, one output layer, and nine hidden layers is designed as demonstrated in figure 4.

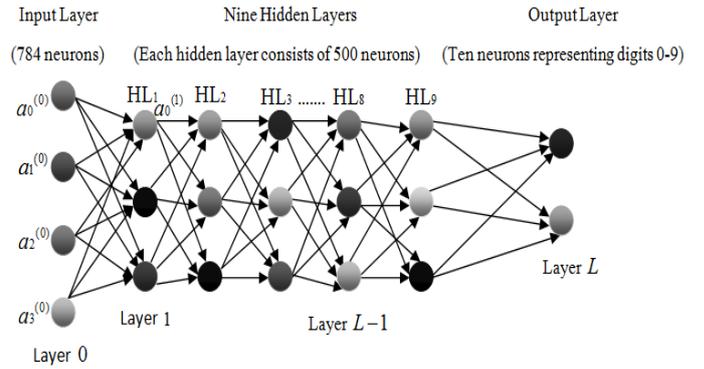

Fig. 4. Stages of an L-layered feed-forward neural network for handwritten digits recognition

The input layer of this network contains 784 neurons as the input data consists of 28 by 28-pixel images of scanned handwritten digits. The input pixels are grayscale values with a 0 for the white pixel, and a perfect 1 for the black pixel and different grayscale pixel values are assigned in between zero to one be subjected to the shadowiness of the images. Each hidden layer contains 500 neurons. The output layer of the system comprises ten neurons to represent digits 0 to 9 respectively. As output neurons are numbered from 0 through 9, the neuron with the highest activation value determines the digit.

To train the network 60000 scanned images of handwritten digits leveled with their correct classifications is used from the MNIST database. All pictures are grayscale and 28 by 28 pixels in size. After completion of training, the network is then tested with 10000 scanned images of digits. Notation x is used to denote training input. As images are 28 by 28 pixels, x is a 784-dimensional vector. The equivalent expected output is indicated by y(x), where y is a ten dimensional vector.

At the outset of the training, all weights and biases in the network are initialized randomly. As the output of the system solely depends on the weight and bias values, the goal of the system is to find appropriate weights and biases so that the output of the system approximates desired output y(x) for all training inputs x. To quantify network performances, a cost function is defined by equation 3 [32].

$$C(w,b) = \frac{1}{2n} \sum_{x} [y(x) - a]^2 \quad ......(3)$$

Here, w = Summation of all the weights in the network
b = All the biases
n = Total number of training samples
a = Actual output

a depends on x, w, and b. C(w,b) =0, precisely when desired output y(x) is roughly tantamount to the actual output, a, for all training samples, x. As all parameters in the network are known except b and w, the job of the training algorithm is to detect correct weights and biases so that C(w,b) = 0. So to lessen the cost C(w,b) to the smallest degree as a function of the weights and biases, the training algorithm has to detect certain weights and biases which cause the cost as small as possible. The algorithm is acknowledged as gradient descent. Gradient descent algorithm utilizes the following equations to set weight and bias values to achieve the global minimum of the cost C(w,b) as demonstrated in figure 5. Again, to enhance the model performance, the notion of stochastic gradient descent was employed.

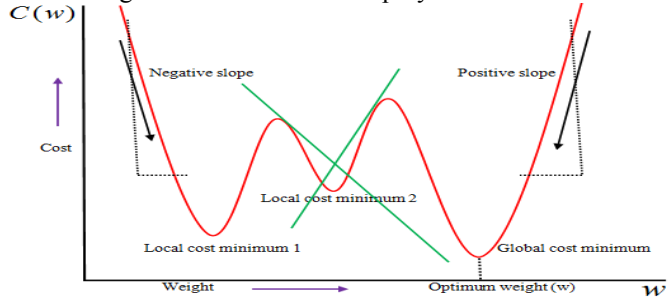

Fig. 5. Diagram Showing the Cost versus Weight Graph In Case Of Neural Network

Gradient Descent [32]:

$$\Delta C \approx \frac{\partial C}{\partial w}\Delta w + \frac{\partial C}{\partial b}\Delta b \quad \ldots\ldots (4)$$

$$\therefore \nabla C \equiv \left(\frac{\partial C}{\partial w}, \frac{\partial C}{\partial b}\right)^T \quad \ldots\ldots (5)$$

$$\nabla C = \begin{bmatrix} \frac{\partial C}{\partial w_{ih}^{(1)}} \\ \frac{\partial C}{\partial b_i^{(1)}} \\ \cdot \\ \cdot \\ \cdot \\ \frac{\partial C}{\partial w_{jk}^{(L)}} \\ \frac{\partial C}{\partial b_j^{(L)}} \end{bmatrix} \quad \ldots\ldots (6)$$

$$w^{new} = w^{old} - \eta \frac{\partial C}{\partial w^{old}} \quad \ldots\ldots (7)$$

$$b^{new} = b^{old} - \eta \frac{\partial C}{\partial b^{old}} \quad \ldots\ldots (8)$$

Stochastic Gradient Descent [32]:

$$C_x = \frac{1}{2}[y(x) - a]^2 \quad \ldots\ldots (9)$$

$$C_x = \frac{1}{n}\sum_x C_x \quad \ldots\ldots (10)$$

$$\therefore \nabla C = \frac{1}{n}\sum_x \nabla C_x \quad \ldots\ldots (11)$$

$$\frac{\sum_{j=1}^{m}\nabla C_{xj}}{m} \approx \frac{\sum_x \nabla C_x}{n} = \nabla C \quad \ldots\ldots (12)$$

$$\nabla C = \frac{1}{m}\sum_{j=1}^{m}\nabla C_{xj} \quad \ldots\ldots (13)$$

$$w^{new} = w^{old} - \frac{\eta}{m}\frac{\partial C_{xj}}{\partial w^{old}} \quad \ldots\ldots (14)$$

$$b^{new} = b^{old} - \frac{\eta}{m}\frac{\partial C_{xj}}{\partial w^{old}} \quad \ldots\ldots (15)$$

Back Propagation [32]:

$$a_0^{(1)} = f(w_{0,0}a_0^{(0)} + w_{0,1}a_1^{(0)} + \ldots + w_{0,n}a_n^{(0)} + b_0) \quad \ldots\ldots (16)$$

$$\sum_{i=0}^{n_{(L)}-1} a_i^{(1)} = f\left(\begin{bmatrix} w_{0,0} & w_{0,1} & . & . & . & w_{0,n} \\ w_{1,0} & w_{1,1} & . & . & . & w_{1,n} \\ . & . & & & & . \\ . & . & & & & . \\ . & . & & & & . \\ w_{i,0} & w_{i,1} & . & . & . & w_{i,n} \end{bmatrix}\begin{bmatrix} a_0^{(0)} \\ a_1^{(0)} \\ . \\ . \\ . \\ a_n^{(0)} \end{bmatrix} + \begin{bmatrix} b_0 \\ b_1 \\ . \\ . \\ . \\ b_n \end{bmatrix}\right) \quad \ldots\ldots (17)$$

$$C = \frac{1}{2n}\sum_{j=0}^{n_{(L)}-1}(y_j - a_j^{(L)})^2 \quad \ldots\ldots (18)$$

$$C_j = \frac{1}{2}(y_j - a_j^{(L)})^2 \quad \ldots\ldots (19)$$

$$a_j^{(L)} = f(z_j^{(L)}) \quad \ldots\ldots (20)$$

$$z_j^{(L)} = \sum_{k=0}^{n_{(L-1)}-1} w_{jk}^{(L)} a_k^{(L-1)} + b_j^{(L)} \quad \ldots\ldots (21)$$

$$\frac{\partial C_j}{\partial w_{jk}^{(L)}} = \frac{\partial z_j^{(L)}}{\partial w_{jk}^{(L)}}\frac{\partial a_j^{(L)}}{\partial z_j^{(L)}}\frac{\partial C_j}{\partial a_j^{(L)}} \quad \ldots\ldots (22)$$

$$\frac{\partial C}{\partial a_j^{(L)}} = \frac{1}{n}\sum_{j=0}^{n_{(L)}-1}(a_j^{(L)} - y_j) \quad \ldots\ldots (23)$$

$$\frac{\partial C_j}{\partial a_j^{(L)}} = (a_j^{(L)} - y_j) \quad \ldots\ldots (24)$$

$$\frac{\partial a_j^{(L)}}{\partial z_j^{(L)}} = f'(z_j^{(L)}) \quad \ldots\ldots (25)$$

$$\frac{\partial z_j^{(L)}}{\partial w_{jk}^{(L)}} = \sum_{k=0}^{n_{(L-1)}-1} a_k^{(L-1)} \quad \ldots\ldots (26)$$

$$\frac{\partial C}{\partial w^{(L)}} = \frac{\partial C}{\partial w_{jk}^{(L)}} = \frac{\partial z_j^{(L)}}{\partial w_{jk}^{(L)}}\frac{\partial a_j^{(L)}}{\partial z_j^{(L)}}\frac{\partial C}{\partial a_j^{(L)}} = \sum_{k=0}^{n_{(L-1)}-1} a_k^{(L-1)} f'(z_j^{(L)}) \frac{1}{n}\sum_{j=0}^{n_{(L)}-1}(a_j^{(L)} - y_j) \quad \ldots\ldots (27)$$

$$\frac{\partial C_j}{\partial w^{(L)}} = \frac{\partial C_j}{\partial w_{jk}^{(L)}} = \frac{\partial z_j^{(L)}}{\partial w_{jk}^{(L)}}\frac{\partial a_j^{(L)}}{\partial z_j^{(L)}}\frac{\partial C_j}{\partial a_j^{(L)}} = \sum_{k=0}^{n_{(L-1)}-1} a_k^{(L-1)} f'(z_j^{(L)})(a_j^{(L)} - y_j) \quad \ldots\ldots (28)$$

$$\frac{\partial C}{\partial w^{(L)}} = \frac{1}{n_{(L)}} \sum_{j=0}^{n_{(L)}-1} \frac{\partial C_j}{\partial w^{(L)}} \ \ldots\ldots\ (29)$$

$$\frac{\partial z_j^{(L)}}{\partial b_j^{(L)}} = 1 \ \ldots\ldots\ (30)$$

$$\frac{\partial C}{\partial b^{(L)}} = \frac{\partial C}{\partial b_j^{(L)}} = \frac{\partial z_j^{(L)}}{\partial b_j^{(L)}} \frac{\partial a_j^{(L)}}{\partial z_j^{(L)}} \frac{\partial C}{\partial a_j^{(L)}} = f'(z_j^{(L)}) \frac{1}{n} \sum_{j=0}^{n_{(L)}-1} (a_j^{(L)} - y_j) \ \ldots\ldots\ (31)$$

$$\frac{\partial C_j}{\partial b^{(L)}} = \frac{\partial C_j}{\partial b_j^{(L)}} = \frac{\partial z_j^{(L)}}{\partial b_j^{(L)}} \frac{\partial a_j^{(L)}}{\partial z_j^{(L)}} \frac{\partial C_j}{\partial a_j^{(L)}} = f'(z_j^{(L)})(a_j^{(L)} - y_j) \ \ldots\ldots\ (32)$$

$$\frac{\partial C}{\partial b^{(L)}} = \frac{1}{n_{(L)}} \sum_{j=0}^{n_{(L)}-1} \frac{\partial C_j}{\partial b^{(L)}} \ \ldots\ldots\ (33)$$

$$\frac{\partial C}{\partial a_k^{(L-1)}} = \sum_{j=0}^{n_{(L)}-1} \frac{\partial z_j^{(L)}}{\partial a_k^{(L-1)}} \frac{\partial a_j^{(L)}}{\partial z_j^{(L)}} \frac{\partial C}{\partial a_j^{(L)}} = \sum_{j=0}^{n_{(L)}-1} \left( \sum_{k=0}^{n_{(L-1)}-1} w_{jk}^{(L)} f'(z_j^{(L)}) \frac{1}{n} \sum_{j=0}^{n_{(L)}-1} (a_j^{(L)} - y_j) \right) \ \ldots\ldots\ (34)$$

$$\frac{\partial C_j}{\partial a_k^{(L-1)}} = \frac{\partial z_j^{(L)}}{\partial a_k^{(L-1)}} \frac{\partial a_j^{(L)}}{\partial z_j^{(L)}} \frac{\partial C_j}{\partial a_j^{(L)}} = \sum_{k=0}^{n_{(L-1)}-1} w_{jk}^{(L)} f'(z_j^{(L)})(a_j^{(L)} - y_j) \ \ldots\ldots\ (35)$$

$$\delta_j^{(L)} = \frac{\partial C}{\partial a_j^{(L)}} \frac{\partial a_j^{(L)}}{\partial z_j^{(L)}} = \frac{1}{n} \sum_{j=0}^{n_{(L)}-1} (a_j^{(L)} - y_j) f'(z_j^{(L)}) \ \ldots\ldots\ (36)$$

$$\delta_j^l = \frac{\partial C}{\partial z_j^l} = \sum_k \frac{\partial C}{\partial z_k^{l+1}} \frac{\partial z_k^{l+1}}{\partial z_j^l} = \sum_k \frac{\partial z_k^{l+1}}{\partial z_j^l} \delta_k^{l+1} = \sum_k w_{kj}^{l+1} \delta_k^{l+1} f'(z_j^l) \ \ldots\ldots\ (37)$$

$$\because z_k^{l+1} = \sum_j w_{kj}^{l+1} a_j^l + b_k^{l+1} = \sum_j w_{kj}^{l+1} f(z_j^l) + b_k^{l+1} \ \ldots\ldots\ (38)$$

$$\therefore \frac{\partial z_k^{l+1}}{\partial z_j^l} = w_{kj}^{l+1} f'(z_j^l) \ \ldots\ldots\ (39)$$

## V. RESULTS AND DISCUSSION

In this paper, we have applied the Neural Network on the MNIST dataset in order of classifying the handwritten digits. We have varied the number of hidden layers and the batch size to observe the variation in the overall classification accuracy.

Figure 6 shows the performance of the Neural Network for different hidden layers accompanied by the variation in batch size. It is observable that in almost all the cases the graph converges after five epochs. However, in the case of 4 hidden layers and 100 batch size, the figure shows a little instability while conversing.

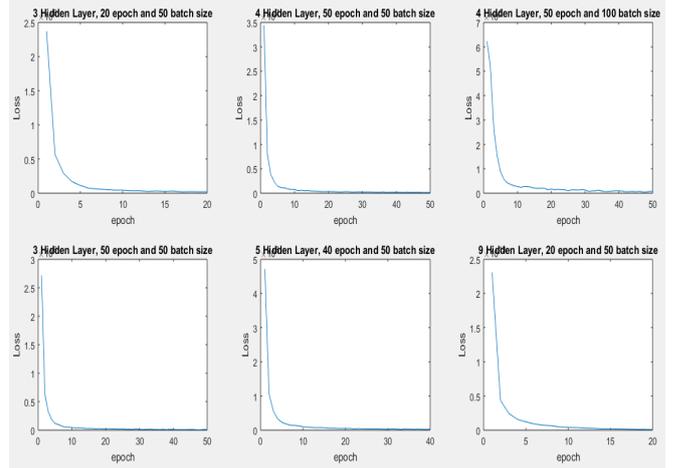

Fig. 6. The photographic representation of the performance comparison of the Neural Network for different hidden layers accompanied by the variation of epoch and batch size

Figure 7 compares the accuracy of the Neural Network corresponding to the number of hidden layers. It shows that the classification accuracy is highest which is 0.9732 for four hidden layers. The second highest point of accuracy is for two hidden layers. On the contrary, the efficiency goes down dramatically with the raise in the number of hidden layers. Though for eight hidden layers the accuracy shows some gain, it decreases again for nine hidden layers. Therefore, according to our observation, four hidden layers show the best accuracy.

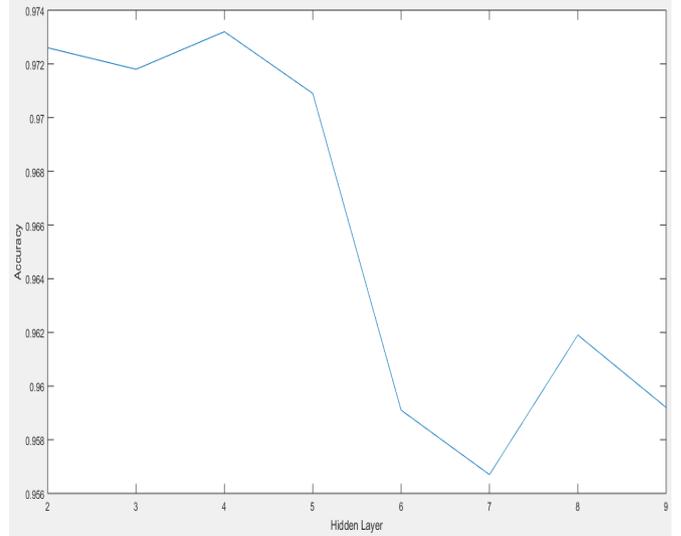

Fig. 7. Comparison of the accuracies of the Neural Network corresponding to the number of hidden layers from hidden layer 2 to hidden layer 9

Table 1 shows the measurement of the respective accuracy according to the number of hidden layers. Though the network demonstrates better performance rate for the higher number of hidden layers, after a while the performance decreases with the rise in the number of hidden layers. It is noted that the computational complexity rises with the increase of the hidden layers as well as the number of neurons.

TABLE I. RESPECTIVE ACCURACIES ACCORDING TO THE VARIATIONS IN THE NEURAL NETWORK

| Number of Hidden Layers | Batch Size | Number of Epoch | Accuracy |
|---|---|---|---|
| 2 | 50 | 50 | 0.9726 |
| 3 | 50 | 20 | 0.9645 |
| 3 | 50 | 50 | 0.9656 |
| 4 | 50 | 20 | 0.9632 |
| 4 | 100 | 20 | 0.9581 |
| 4 | 50 | 50 | 0.9732 |
| 4 | 100 | 50 | 0.9656 |
| 5 | 50 | 40 | 0.9709 |
| 6 | 50 | 20 | 0.9591 |
| 7 | 50 | 20 | 0.9567 |
| 8 | 50 | 20 | 0.9619 |
| 9 | 50 | 20 | 0.9592 |

## VI. CONCLUSION

In recent time, the demand for artificial intelligence is increasing day by day leaving the machine to handle most of the works of the human. Researches on upgrading the response of machine core as smooth as the performance of the human brain are going on. The performance of the machine can be naturalized by the development of the accuracy in the performance of the neural network with the moderate cost level. In this paper, the efficiency in the performance of Neural Network is demonstrated by varying the hidden layers with non-identical batch sizes. The accuracies were observed mostly for three and four hidden layers by changing the number of the epoch. For 50 epochs the highest accuracy was found for 4 hidden layers whereas the best performance among the 20 epochs was illustrated for 3 hidden layers. The maximum accuracy in the performance was found 97.32% among all the observations for four hidden layers with 50 batch sizes though the cost level is required to be kept minimized. This kind of higher accuracy will collaborate to promote the performance of machine more effectively in random digit recognition. On the contrary, further precision in the performance of the neural network with the decrease in the cost function with better image resolution can be attained by Convolutional Neural Network researches. Our future plan is to implement the analysis for other datasets which would lead us to get an overview of the impact of layers in the neural network over the dimensions of the datasets.